\documentclass{article}

 \usepackage[preprint]{neurips_2026}

\usepackage[utf8]{inputenc} 
\usepackage[T1]{fontenc}    
\usepackage{hyperref}       
\usepackage{url}            
\usepackage{booktabs}       
\usepackage{amsfonts}       
\usepackage{nicefrac}       
\usepackage{microtype}      
\usepackage{xcolor}         
\usepackage{amsmath}
\usepackage[ruled,vlined]{algorithm2e}
\usepackage{graphicx}
\usepackage{subcaption}
\title{GES-TSP: Graph Edge Sparsification for TSP}

\author{
  Tianfeng Chen \\
  School of Mathematics and Statistics \\
  Lanzhou Universtiy\\
  \texttt{chentf2025@lzu.edu.cn}
  \And
  Xianyue Li* \\
  School of Mathematics and Statistics \\
  Lanzhou Universtiy\\
  \texttt{lixianyue@lzu.edu.cn} 
}
\begin{document}
\maketitle
\begin{abstract}
  Solving large-scale instances of the Traveling Salesman Problem (TSP) exactly is computationally expensive. Researchers often employ graph sparsification methods to improve computational efficiency. Traditional sparsification methods typically rely on fixed heuristics and fail to fully exploit instance-specific structural information. In this paper, we propose Graph Edge Sparsification (GES), a learning-based sparsification approach for Euclidean TSP. By incorporating geometric structural information and combinatorial optimization technology, our proposed method adaptively generates a sparsification graph for different instances, significantly reducing the graph size and accelerating the solving process. Experimental results demonstrate that our sparsification method can prune up to 95\% of edges on the MATILDA dataset, while keeping the solution gap within 1\% of the optimal value. Moreover, our approach exhibits strong generalization capability on the TSPLIB benchmark.In some large-scale instances, the pruning rate exceeds 99\%, while the optimality gap remains below 1\%.
\end{abstract}

\section{Introduction}

The Traveling Salesman Problem (TSP) is a classical NP-hard combinatorial optimization problem \cite{junger1995traveling}. Exact methods are computationally expensive and often require lots of time to solve large-scale instances. As a fundamental benchmark problem in combinatorial optimization, the Euclidean TSP has been extensively studied due to its broad applications in transportation planning, circuit design, and routing systems. Improving the efficiency of solving Euclidean TSP instances is therefore of both theoretical and practical significance. In particular, Euclidean TSP instances are typically formulated on complete graphs, where the number of edges grows quadratically with the number of nodes, leading to significant computational overhead in exact or high-quality approximate solvers.

Graph sparsification \cite{spielman2011spectral} is a widely used strategy to reduce the computational complexity of TSP instances by restricting the search space to a subset of candidate edges. Traditional approaches are mainly based on geometric heuristics, which construct sparse graphs without exploiting instance-specific information.

One of the most commonly used sparsification techniques is the $k$-nearest neighbor (KNN) graph \cite{de2008computational}, where each node is connected to its $k$ closest neighbors in terms of Euclidean distance. Another widely adopted sparsification method is Delaunay triangulation \cite{xu2020delaunay}, which constructs a planar graph by maximizing the minimum angle of all triangles. Delaunay graphs possess strong geometric properties and are known to contain many edges of optimal Euclidean TSP tours in practice. However, although it provides a more structured sparsification compared to KNN graphs, it still includes redundant edges or miss problem-specific structures in certain distributions.

In recent years, learning-based end-to-end approaches have been proposed for TSP \cite{bresson2021transformer,joshi2019efficient,vinyals2015pointer,bello2016neural,kwon2020pomo}. These methods directly learn to construct tours from data, typically using sequence models or attention-based architectures. However, they often exhibit limited generalization capability and involve complex model architectures with a large number of parameters.

Recent studies have explored learning-based approaches for graph sparsification. Instead of directly constructing tours, these methods aim to identify a subset of promising edges that are likely to appear in high-quality solutions. In particular, Graph Neural Networks (GNNs) can leverage node and edge features to predict edge importance, enabling the construction of sparse graphs that significantly reduce computational complexity while preserving solution quality.

\section{Related works}

Fitzpatrick et al.  \cite{fitzpatrick2021learning} propose a learning-based method that reformulates the TSP sparsification problem as a binary classification task, with the aim of retaining edges that are likely to appear in the optimal tour. They extract a variety of edge features to characterize the structural and geometric properties of the graph. 
Based on these features, they use traditional machine learning models, such as logistic regression and support vector machines, to predict the importance of edges, typically formulated as a binary classification problem. 
Xin et al. \cite{xin2021neurolkh} propose an Sparse Graph Network (SGN) framework to construct a sparse candidate edge set  to improve the Lin-Kernighan-Helsgaun (LKH) heuristic. They use node coordinates as node features and Euclidean distances as edge features. 
Based on these inputs, they use proposed SGN to predict edge scores. 
Afterwards, for each node, only the top-$k$ edges with the highest scores are retained to construct a candidate edge set.
Tian et al. \cite{tian2024combhelper} also use GNNs for graph sparsification, but their approach mainly focuses on the Vertex Cover and Maximum Independent Set problems. However, these methods either suffer from limited generalization capability or fail to fully exploit structural information of the graph.

In this paper, we propose GES, a learning-based sparsification approach that effectively captures structural information and instance-specific graph data, and integrates combinatorial optimization techniques to improve both efficiency and solution quality.

\section{Preliminaries}

\subsection{TSP and Graph Sparsification}

The TSP seeks a shortest tour that visits each vertex exactly once and returns to the starting point. Given a weighted graph $G=(V,E)$, the goal is to find a Hamiltonian cycle with minimum total cost, where each edge $(i,j)$ is associated with a cost $c_{ij}$.

In this work, we focus on the Euclidean TSP (ETSP), where each vertex $i \in V$ is embedded in a 2D coordinate $p_i=(x_i,y_i)$, and edge costs are defined as Euclidean distances between points.

Since the Euclidean TSP is defined on a complete graph, the number of edges increases rapidly as the number of vertices grows. 
As a result, solving the problem to optimality using exact solvers can be computationally expensive, especially for large-scale instances. 
This motivates the use of graph sparsification techniques to reduce the problem size while preserving solution quality.

Graph sparsification \cite{hashemi2024comprehensive} selects existing edges from the graph $G$, and outputs $G'$=$(V,E')$, where $E'$ is the subset of $E$. Our goal is to reduce the number of edges in the graph as much as possible while preserving high-quality TSP solutions.

\subsection{GNNs}

Graph Neural Networks (GNNs) are widely used for learning on graph-structured data \cite{scarselli2008graph,hu2020open}. Typical GNNs update node representations through iterative message passing, where each node aggregates information from its neighbors. 

Among various GNNs architectures, this paper adopts the Graph Attention Network (GAT) \cite{velickovic2018graph}, which uses an attention mechanism to learn the importance of neighboring nodes adaptively. For node $v$ and its neighbor $u \in \mathcal{N}(v)$, the attention coefficient is computed as:
\begin{equation}
e_{vu} = \text{LeakyReLU}\left( \mathbf{a}^{\top} [\mathbf{W} \mathbf{h}_v \| \mathbf{W} \mathbf{h}_u \| \mathbf{W}_e \mathbf{r}_{vu}] \right),
\end{equation}
where $\mathbf{h}_v^{(l)}$ denotes the embedding of node $v$, $\mathcal{N}(v)$ denotes its neighbor set, $\mathbf{W}$ and $\mathbf{W}_e$ are learnable weight matrices, $\mathbf{a}$ is the attention vector, $\mathbf{r}_{vu}$ denotes edge features, and $\|$ represents concatenation. The normalized attention coefficients are then used to aggregate neighbor information:
\begin{equation}
\mathbf{h}_v' = \sigma\left( \sum_{u \in \mathcal{N}(v)} \alpha_{vu} \mathbf{W} \mathbf{h}_u \right).
\end{equation}

Compared with traditional GNNs, GAT can better capture the relative importance of different neighbors, making it particularly suitable for graph optimization problems such as the TSP.

\subsection{Delaunay Triangulation}
The Delaunay triangulation is a fundamental structure in computational geometry that provides a sparse graph representation for a set of points in the plane. 
Given a set of points in $\mathbb{R}^2$, the Delaunay triangulation constructs a triangulation such that no point lies inside the circumcircle of any triangle. 

An important property of the Delaunay triangulation is that it preserves proximity relationships between points and contains many edges that are likely to appear in the optimal Euclidean TSP tour \cite{xu2020delaunay}. 
Moreover, the number of edges in the Delaunay triangulation is linear in the number of vertices, i.e., $\mathcal{O}(n)$, which is significantly smaller than the $\mathcal{O}(n^2)$ edges in the complete graph.

\subsection{Christofides algorithm}
The Christofides algorithm \cite{christofides2022worst} is a classical approximation algorithm for the TSP with a worst-case approximation ratio of $3/2$. It first computes a Minimum Spanning Tree (MST), which is a connected subgraph spanning all vertices with minimum total edge weight. Then, a minimum-weight perfect matching is constructed on the odd-degree vertices of the MST, where each selected vertex is matched exactly once with minimum total matching cost. By combining the MST and matching edges, an Eulerian graph is obtained, i.e., a graph containing a closed trail that traverses every edge exactly once. Finally, repeated vertices in the Eulerian tour are shortcut to produce a Hamiltonian cycle. The resulting tour satisfies $c(C)\le \frac{3}{2}c(\mathrm{OPT})$, where $c(\mathrm{OPT})$ denotes the optimal tour length.

\section{GES-TSP}
Common approaches for solving the TSP include exact methods, heuristic methods, and approximation algorithms. 
Exact methods typically rely on optimization solvers such as CPLEX and SCIP \cite{achterberg2009scip} to obtain optimal solutions, but they are computationally expensive and do not scale well to large instances. 
Compared to exact methods, heuristic approaches are more scalable and can efficiently produce high-quality solutions for large-scale instances, among which the LKH algorithm \cite{helsgaun2015solving} is one of the most effective.
Approximation algorithms provide theoretical performance guarantees and can compute solutions within a bounded ratio of the optimum in polynomial time.

Most methods of TSP operate on the complete graph, leading to significant computational overhead due to the $\mathcal{O}(n^2)$ number of edges. 
This motivates the use of graph sparsification techniques to reduce the problem size. 
In particular, our GES-TSP offers a promising data-driven approach to identify relevant edges and enable efficient TSP solving.
Figure 1 shows the overview of the GES-TSP and solving framework.

\begin{figure}[htbp]
    \centering
    \includegraphics[width=0.7\linewidth]{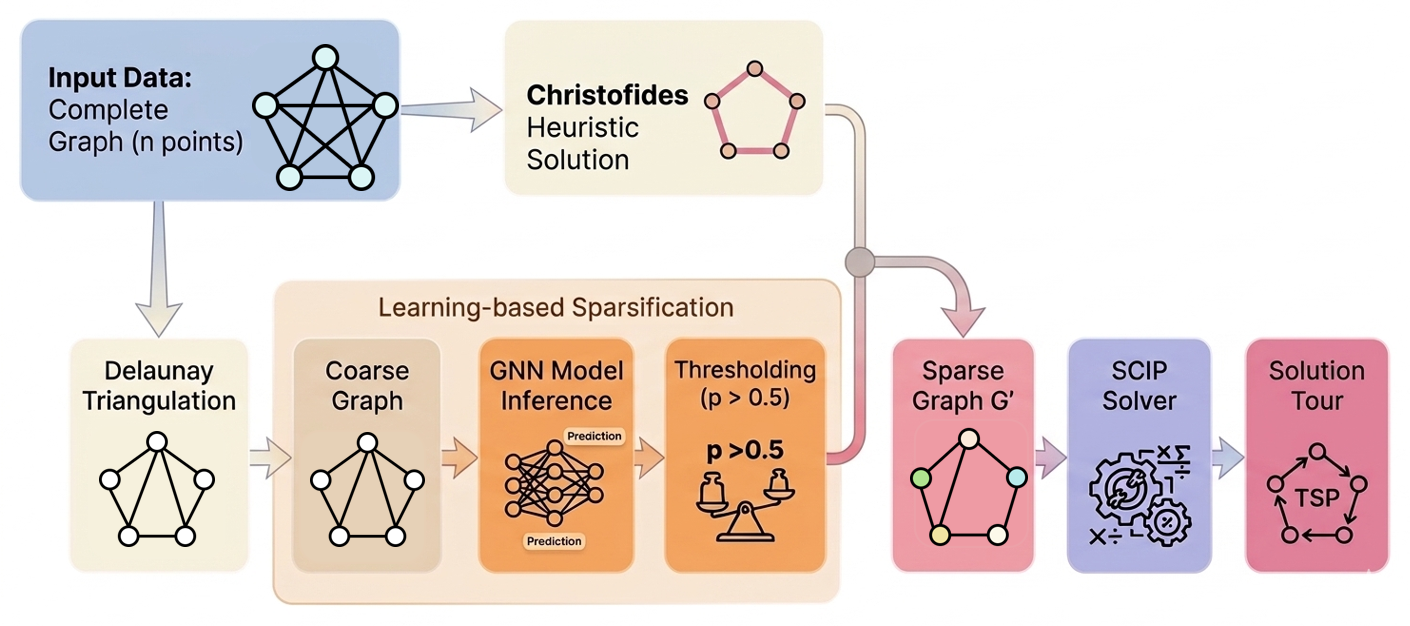}
    \caption{Overview of the GES-TSP and Solving Framework}
    \label{fig:Framework}
\end{figure}

\subsection{Coarse Graph}
To reduce the computational burden of solving the TSP and training GNNs models on a complete graph, we first perform a coarse graph sparsification based on geometric structures.

Specifically, given a set of points in the Euclidean plane, we construct the Delaunay triangulation, which provides a sparse graph that preserves important proximity relationships between vertices. 
The Delaunay graph contains $\mathcal{O}(n)$ edges and is known to retain many edges that are likely to appear in high-quality TSP tours.

By leveraging the Delaunay triangulation, we obtain an initial sparse graph that significantly reduces the number of edges while preserving local geometric structure. 
This coarse graph serves as a strong candidate edge set and provides a more efficient input for subsequent learning-based refinement using GNNs.

\subsection{Features Construction}
After obtaining the coarse graph, we construct feature representations for both nodes and edges. Node features are defined as normalized coordinates, while edge features are carefully designed to capture structural information, since TSP primarily depends on edge selection.

In the Euclidean TSP, the cost of an edge is determined by the Euclidean distance between two nodes. We directly use this distance as a fundamental edge feature. In optimal tours, shorter edges are significantly more likely to be included, while longer edges are rarely selected. However, distance alone is insufficient. It does not capture local structure and may not work well in regions with different node densities. This limitation motivates the incorporation of additional features.

We choose the KNN feature as the second edge feature. For each node $i$, let $\mathcal{N}_k(i)$ denote the set of its $k$ nearest neighbors. For an edge $(i,j)$, we define a binary variable:
\begin{equation}
\text{KNN}_{ij} =
\begin{cases}
1, & \text{if } j \in \mathcal{N}_k(i) \text{ or } i \in \mathcal{N}_k(j), \\
0, & \text{otherwise}.
\end{cases}
\end{equation}

This feature shows whether two nodes are close and provides local structure information. It helps the model handle regions with different densities. 

While the KNN feature captures whether two nodes are locally close, it only provides a binary signal and cannot reflect the relative quality of edges. To address this limitation, we introduce stronger local features. For an edge $(i,j)$, it is defined as
\begin{equation}
Q_{ij} = \frac{1 + d_{ij}}{1 + \min_{k} d_{ik}}.
\end{equation}

Fitzpatrick et al.  \cite{fitzpatrick2021learning} constructs six local features. In this work, we simplify them into one feature. This feature measures how good an edge is compared with the shortest edge of node $i$. A smaller value indicates a more preferred connection. Compared with the KNN feature, the Q-value provides a finer-grained measure of local edge quality \cite{sun2020generalization}. It does not only indicate whether two nodes are close, but also reflects how much worse an edge is relative to the best local choice. This makes it more informative for edge selection.

Finally, we also need global features to fully describe the structure of a graph. Fitzpatrick et al.
\cite{fitzpatrick2021learning} propose IMST features. We iteratively compute the MST of the current graph, remove its edges, and assign each selected edge a weight of $1/k$ at iteration $k$. The process runs for at most $\lceil \log n \rceil$ iterations or until the graph becomes disconnected. Finally, all edge weights are normalized over the accumulated edge set. In our method, the IMST feature is further normalized to ensure numerical stability across different graph sizes.
The iteration depth $R = \lceil \log_2 n \rceil$ depends on graph size, causing the accumulated IMST weights to grow with $n$. Without normalization, this leads to scale-inconsistent features and unstable GNNs training across graphs of different sizes. Therefore, we normalize the IMST weights to make the feature scale-invariant and reflect only the relative importance of edges.

Therefore, each edge is represented by a four-dimensional feature vector.
\begin{equation}
e_{ij}=[d_{ij},\text{KNN}_{ij}, \text{IMST}_{ij}, \text{Q}_{ij}]
\end{equation}

\subsection{GNN Model Inference}

We use GAT to predict the importance of edges in the TSP graph. First, each node coordinate is mapped into a hidden representation using a linear transformation. Then, we apply a GAT-based message passing network. In each layer, node representations are updated by aggregating information from neighboring nodes, while edge features are incorporated during the message passing process. ReLU activation is applied after each layer. 

After message passing, we construct an edge representation by concatenating the source node embedding, the target node embedding, and the corresponding edge features. This representation is then fed into a multi-layer perceptron (MLP), which outputs a score for each edge. Higher scores indicate a higher likelihood of being included in the optimal tour.
The model is trained as a binary edge classification problem using the binary cross-entropy loss with logits. Figure 2 shows the overall training pipeline of the proposed method.

\begin{figure}[htbp]
    \centering
    \includegraphics[width=0.8\linewidth]{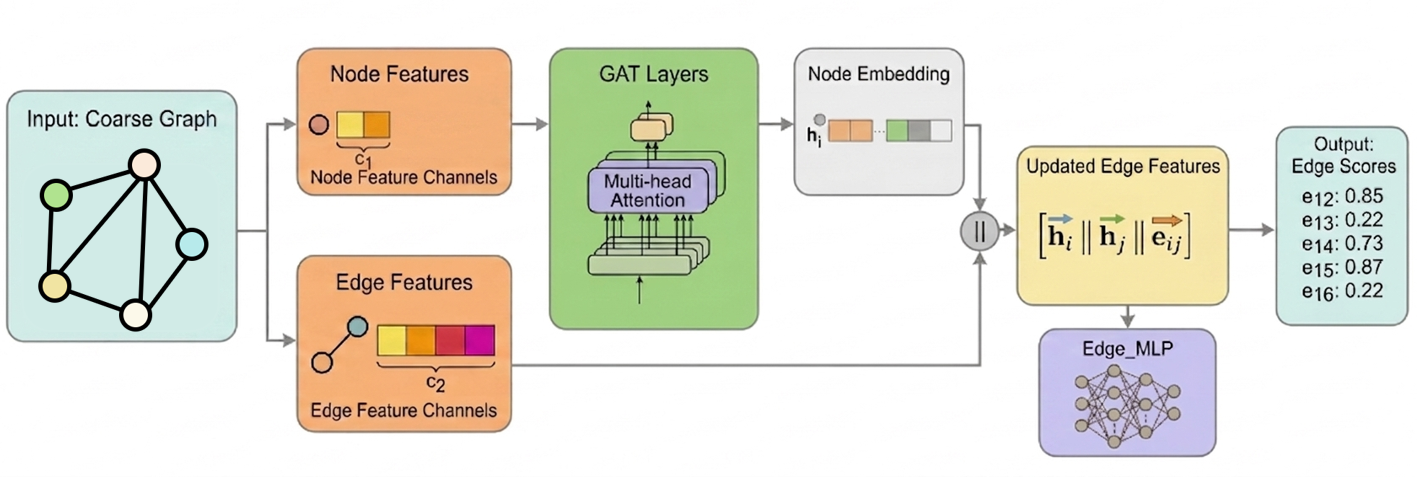}
    \caption{Architecture of the Proposed GNNs Model}
    \label{fig:GNN}
\end{figure}

\subsection{Sparse Graph}
Now we obtain the predicted scores for all edges in the full graph. These scores reflect how likely each edge is to be part of the optimal tour.
We then use these scores to build a sparse graph. A threshold is applied to filter edges: edges with scores higher than the threshold are kept, while the others are removed. In this way, we keep only a small set of candidate edges.
This step greatly reduces the number of edges in the graph. At the same time, most important edges are still kept, since they usually receive higher scores from the model. Therefore, the graph becomes much smaller but still contains the key structure of the original problem.

To ensure that a feasible solution always exists, we further include the edges from the Christofides algorithm solution. These edges guarantee that the sparse graph remains connected and contains at least one valid tour.

The sparse graph is then passed to an optimization solver to compute the final TSP solution. Since the graph is smaller, the solver can run faster and use less memory. This makes it possible to handle larger problem instances more efficiently. Figure 3 shows a sparse graph for a TSP instance (CLKhard 070). Orange edges appear only in the sparse-graph solution, blue edges appear only in the global optimum. Red edges are shared by both tours.

\begin{figure}[htbp]
    \centering
    \includegraphics[width=0.4\linewidth]{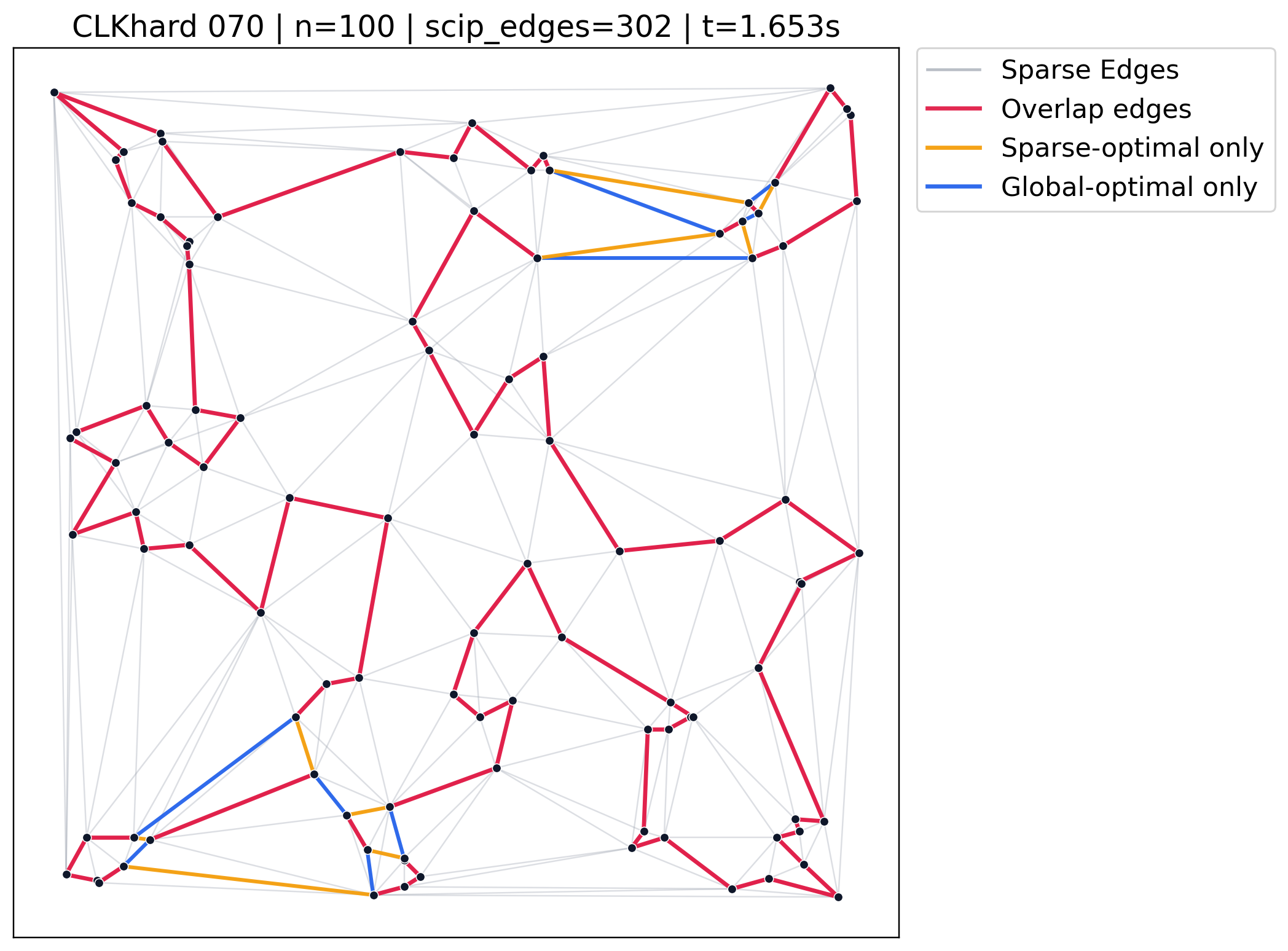}
    \caption{Sparse Graph for a TSP Instance}
    \label{Sparse Graph}
\end{figure}

\section{Experiments and Results}
All experiments are implemented in Python 3.12, and the optimization is performed using SCIP. All computations were performed on a Dell laptop running Windows 11 with 15.8 GB of RAM, an Intel® Core™ i7-10870H 2.60GHz CPU and an NVIDIA GeForce RTX 3060 Laptop GPU. We use a GATv2-based graph neural network to perform binary classification on candidate TSP edges.  The model consists of 3 GATv2Conv layers with hidden dimension 64 and 4 attention heads. Edge prediction is performed by concatenating node embeddings and edge features, followed by an MLP with dropout rate 0.1. The network is trained using the Adam optimizer with learning rate 0.001, batch size 4, and 50 training epochs. To address class imbalance, BCEWithLogitsLoss with positive class weight 30 is adopted. 

\subsection{Comparative study on TSP}

We primarily evaluate the performance of GES-TSP on the MATILDA and TSPLIB benchmarks, we consider instances ranging from ch130 to pr2392, since optimal solution files are not available for larger instances beyond pr2392. and compare it with two representative baseline methods from the existing literature. Xin et al. \cite{xin2021neurolkh} propose an SGN-based framework, which uses GNNs to score edges and constructs a sparse candidate set by retaining the top-$k$ highest-scoring edges for each node. Fitzpatrick et al.  \cite{fitzpatrick2021learning} formulate graph sparsification as a binary edge classification problem and employ classical machine learning models to construct a sparse graph.

The MATILDA dataset consists of seven categories, each containing 190 instances of 100-node 2D Euclidean TSP problems. We use one third of the CLKhard and LKCChard instances for training, and the rest for validation and testing (the same as in \cite{fitzpatrick2021learning}).

\begin{table}[htbp]
\centering
\caption{Optimality Gap Comparison of TSP Solving on the MATILDA }
\label{tab:Optimality Ratio}
\footnotesize
\begin{tabular*}{\textwidth}{@{\extracolsep{\fill}}lccc} 
\toprule
Statistic & SGN & Fitzpatrick & GES  \\
\midrule
CLKeasy & 0.43\% & \textbf{0.00\%} & \textbf{0.00\%} \\
CLKhard & 1.49\% & 0.18\% & \textbf{0.14\%} \\
LKCCeasy & 0.41\% & \textbf{0.00\%} & 0.03\% \\
LKCChard & 1.43\% & 0.38\% & \textbf{0.20\%} \\
easyCLK-hardLKCC & 0.68\% & 0.17\% & \textbf{0.10\%} \\
hardCLK-easyLKCC & 0.72\% & 0.13\% & \textbf{0.03\%} \\
random & 0.76\% & 0.25\% & \textbf{0.16\%} \\
\bottomrule
\end{tabular*}
\end{table}

Table 1 presents the optimality gap of the three methods with respect to the optimal solutions. Except for CLKeasy and LKCCeasy, GES consistently outperforms the other methods by achieving the lowest optimality gap across all remaining datasets. On more challenging datasets such as CLKhard and LKCChard, our method significantly outperforms SGN and consistently improves upon Fitzpatrick. On simpler datasets such as CLKeasy and LKCCeasy, all methods achieve near-optimal solutions, leaving limited room for improvement.

\begin{figure}[htbp]
    \centering
    \includegraphics[width=0.8\linewidth]{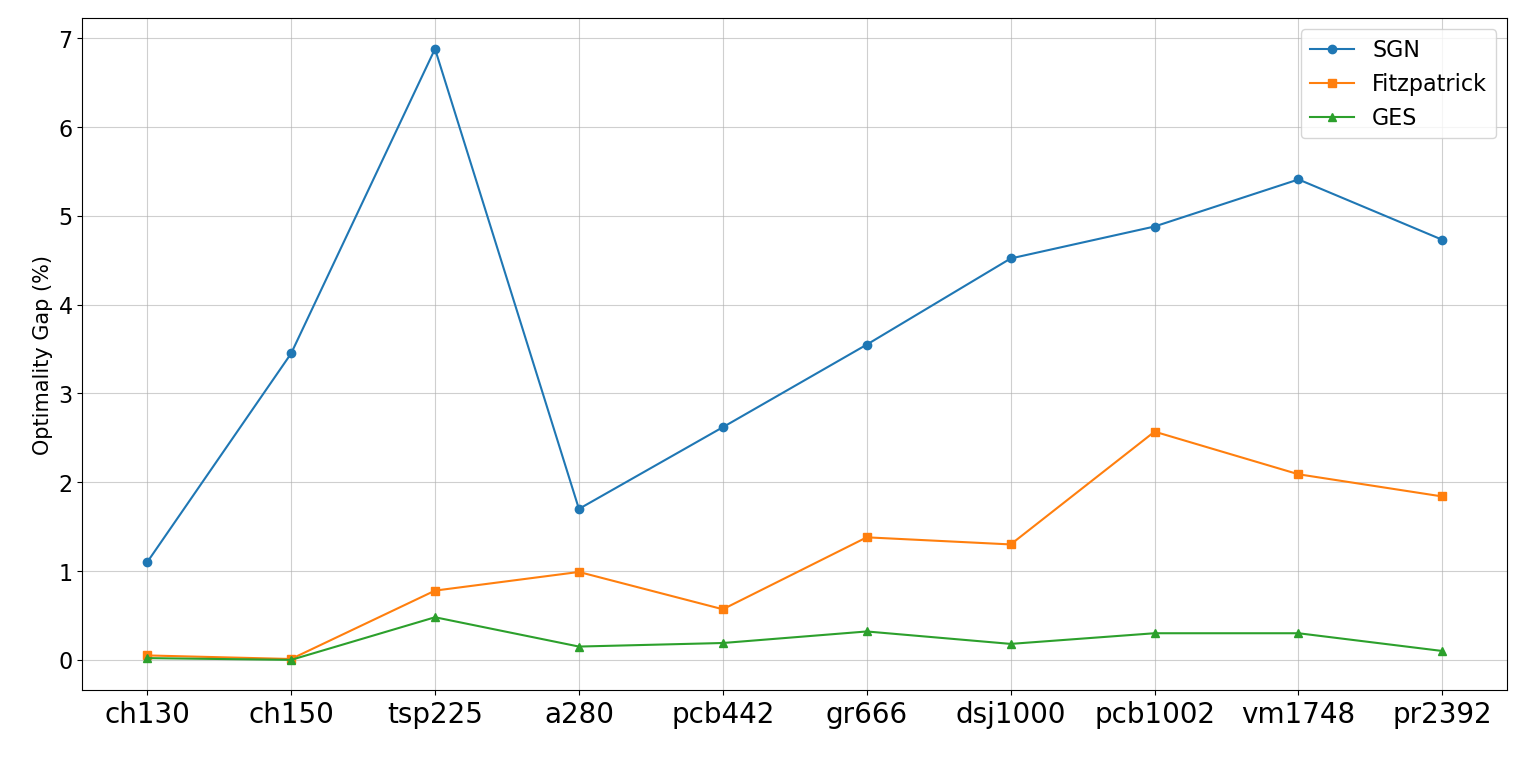}
    \caption{Optimality Gap Comparison of TSP Solving on the TSPLIB}
    \label{fig:charttsplib}
\end{figure}

Figure 4 shows the optimality gap of different methods on TSPLIB instances. 
Our method consistently achieves the lowest optimality gap across all instances.
Compared with SGN, our method significantly reduces the gap, especially on larger and more challenging instances such as tsp225 and vm1748, etc.
Compared with Fitzpatrick, our method still demonstrates consistent improvements, particularly on large-scale instances including pcb1002, vm1748, and pr2392.
Moreover, the performance advantage becomes more pronounced as the problem size increases, indicating strong scalability. 
Overall, our method maintains a stable and near-optimal performance across all instances, demonstrating excellent generalization capability.

\begin{table}[htbp]
\centering
\caption{Pruning Rate Comparison of TSP Solving on the MATILDA }
\label{tab:Pruning Rate}
\footnotesize
\begin{tabular*}{\textwidth}{@{\extracolsep{\fill}}lccc} 
\toprule
Statistic & SGN & Fitzpatrick & GES  \\
\midrule
CLKeasy & 92.81\% & 90.38\% & \textbf{94.24\%} \\
CLKhard & 92.13\% & 89.02\% & \textbf{93.93\%} \\
LKCCeasy & 93.87\% & 91.08\% & \textbf{94.22\%} \\
LKCChard & 92.45\% & 88.29\% & \textbf{94.04\%} \\
easyCLK-hardLKCC & 93.71\% & 88.87\% & \textbf{94.08\%} \\
hardCLK-easyLKCC & 93.35\% & 90.35\% & \textbf{94.16\%} \\
random & 93.18\% & 89.44\% & \textbf{94.02\%} \\
\bottomrule
\end{tabular*}
\end{table}

Table 2 reports the pruning rate of different methods on the MATILDA dataset. 
GES consistently achieves the highest pruning rate across all datasets.
Compared with SGN and Fitzpatrick, GES significantly increases the pruning rate while maintaining stability across different instance types. 
In particular, the improvement over Fitzpatrick is significant on harder datasets such as CLKhard and LKCChard.
Although SGN achieves a relatively high pruning rate, Table 1 shows that it suffers from a larger optimality gap.

We further evaluate the pruning rate on TSPLIB instances to assess the generalization ability of our method. 
As shown in Figure 5, GES consistently achieves the highest pruning rate across all TSPLIB instances.
Compared with SGN and Fitzpatrick, our method maintains a clear advantage, especially on large-scale instances such as pcb1002, vm1748, and pr2392, where the pruning rate exceeds 99\%. 
In contrast, although SGN also achieves relatively high pruning rates, it remains consistently lower than GES, while Fitzpatrick shows a more noticeable gap.
Moreover, the pruning rate of GES remains highly stable across instances of varying scales, demonstrating strong robustness and generalization beyond the training distribution.

\begin{figure}[htbp]
    \centering
    \includegraphics[width=0.8\linewidth]{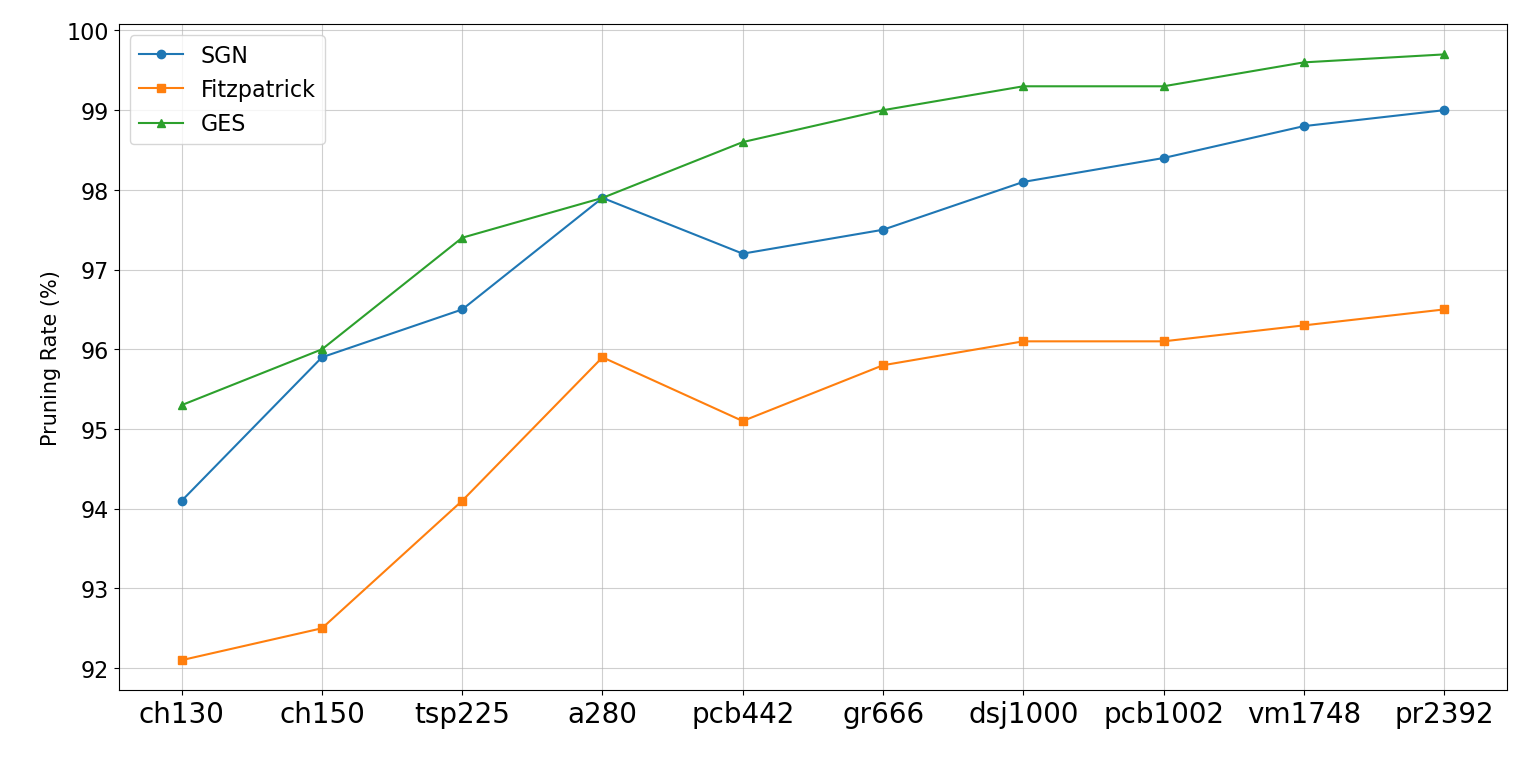}
    \caption{Pruning Rate Comparison of TSP Solving on the TSPLIB}
    \label{fig:pruning}
\end{figure}

\begin{figure}[htbp]
    \centering    \includegraphics[width=0.8\linewidth]{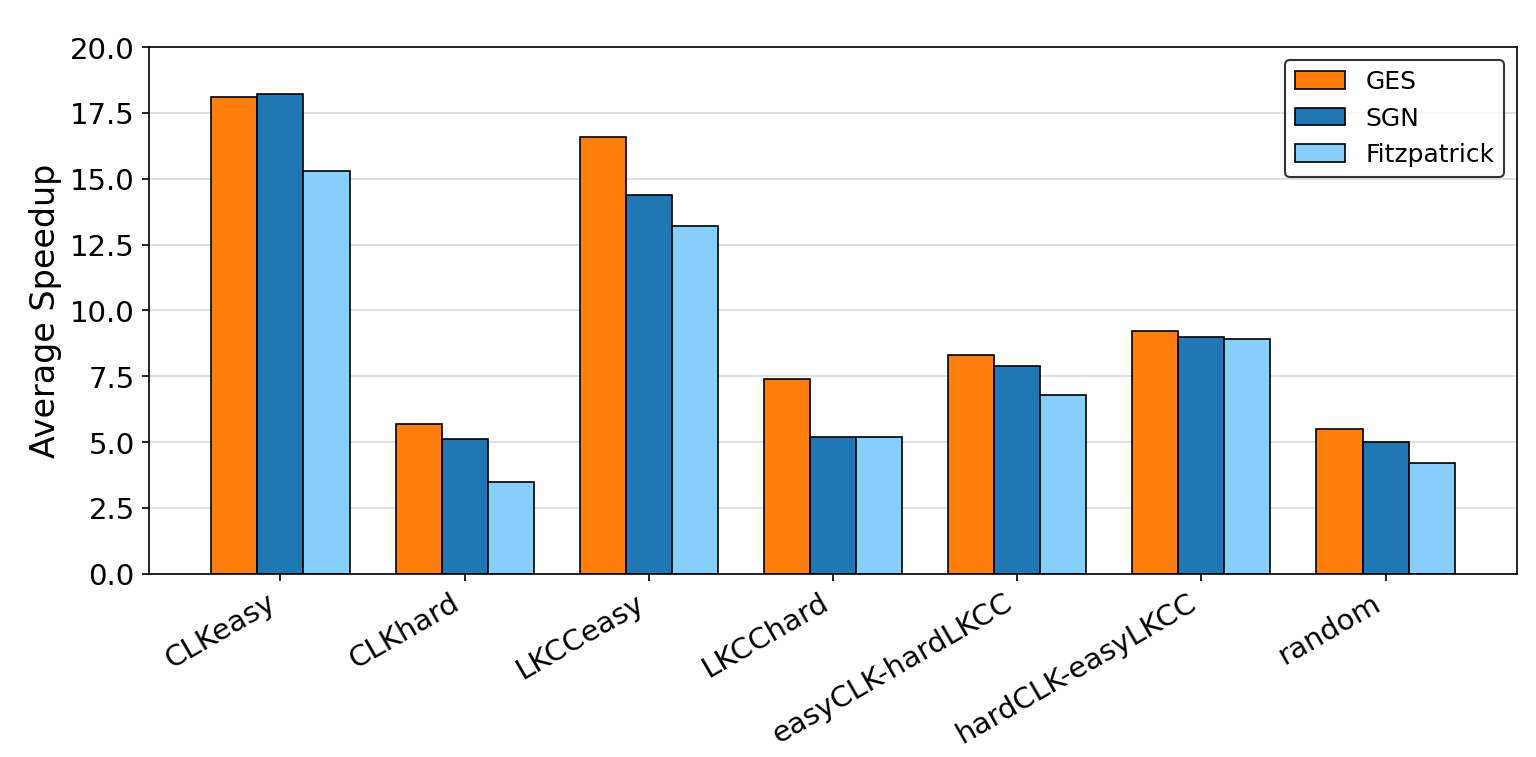}
    \caption{Average speed-up on the MATILDA}
    \label{fig:placeholder}
\end{figure}

After showing that our method achieves superior solution quality and higher pruning rates, we further analyze the solving time of the three methods. Figure 6 shows the average speedup of the three methods on the MATILDA dataset compared to directly solving the original graph. It can be observed that GES consistently achieves the best speedup performance, followed by SGN, while Fitzpatrick performs the worst. This ranking is highly consistent with the pruning rate patterns we obtained.

Figure 7 illustrates the distribution of edge prediction scores generated by the GNNs model. 
Kernel Density Estimation (KDE) is employed to provide a smooth approximation of the underlying distributions. 
The blue curve denotes the distribution over all edges, the red curve corresponds to positive edges (i.e., edges belonging to the optimal tour), and the green curve represents negative edges.

It can be observed that most edges receive low scores (below 0.2), while a relatively small proportion of edges are assigned high scores (above 0.8). 
This aligns with the class imbalance in the complete TSP graph, where the number of negative edges far exceeds that of positive ones.
Moreover, positive edges are predominantly concentrated near 1, while negative edges cluster around 0, indicating that the model exhibits strong discriminative capability in identifying edges that are likely to belong to the optimal tour.

\begin{figure}[htbp]
    \centering
    \includegraphics[width=0.6\linewidth]{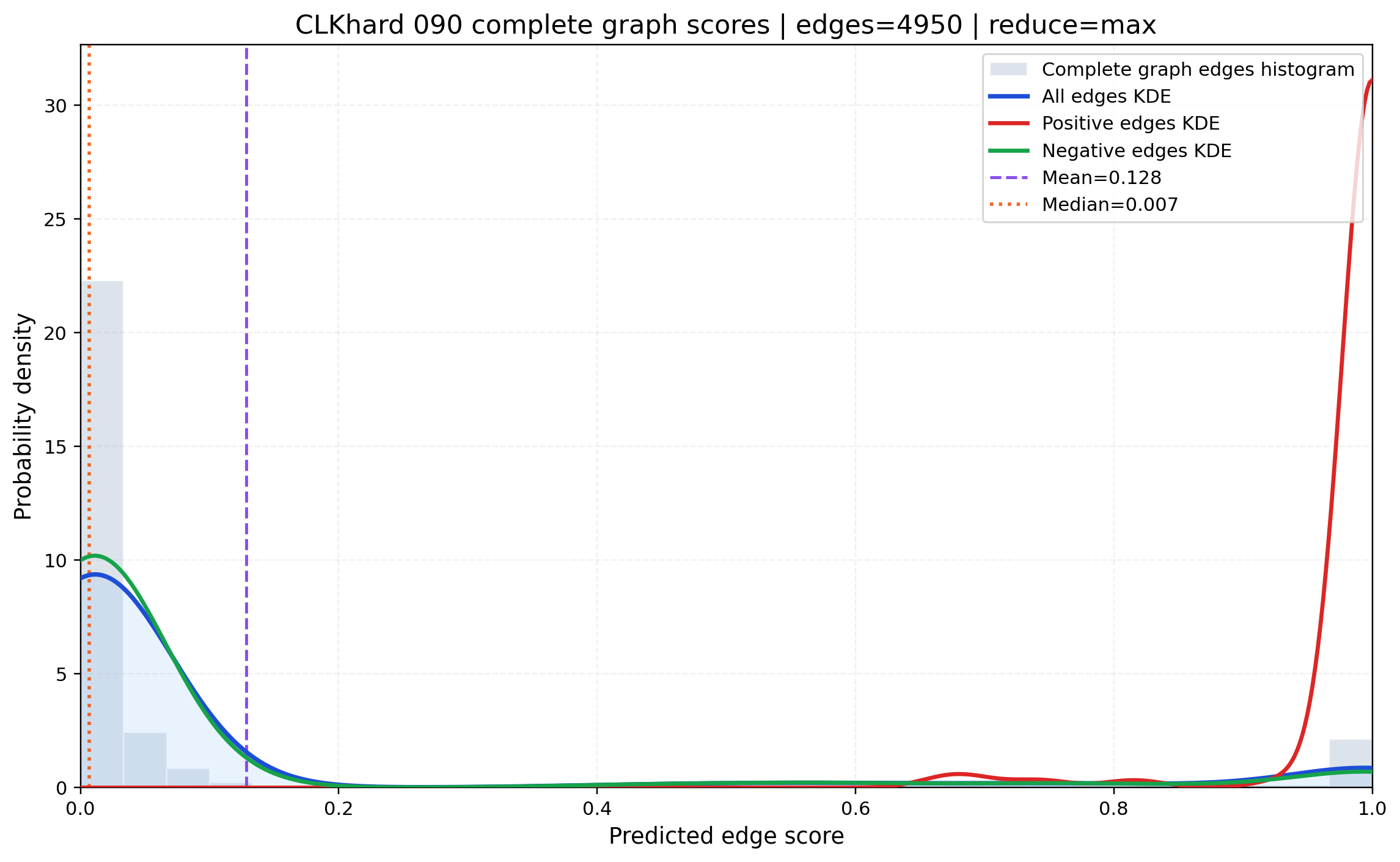}
    \caption{Distribution of Edge Prediction Scores}
    \label{fig:placeholder}
\end{figure}

\subsection{Ablation Study}
We first conduct a comparative study on different coarse filtering strategies. Table 3 shows the differences in the pruning rate when removing the Delaunay triangulation and replacing it with a KNN-based coarse filtering strategy. It can be observed that, without the Delaunay triangulation, the overall pruning rate decreases by approximately 5\%. When the Delaunay triangulation is replaced with a KNN-based coarse filtering strategy, as adopted in SGN, the pruning rate decreases by approximately 3\%. 

\begin{table}[htbp]
\centering
\caption{Pruning Rate Comparison with Different Coarse Filtering Strategies}
\label{tab:cf}
\footnotesize
\begin{tabular*}{\textwidth}{@{\extracolsep{\fill}}lccc} 
\toprule
Statistic & GES & w/o Delaunay Triangulation & GES (KNN-based)  \\
\midrule
CLKeasy  & \textbf{94.24\%} & 89.75\% & 91.21\%  \\
CLKhard & \textbf{93.93\%} & 89.19\% & 90.39\%  \\
LKCCeasy & \textbf{94.22\%} & 89.67\% & 92.18\% \\
LKCChard & \textbf{94.04\%} & 89.81\% & 91.45\% \\
easyCLK-hardLKCC & \textbf{94.08\%} & 90.16\% & 92.05\% \\
hardCLK-easyLKCC & \textbf{94.08\%} & 89.53\% & 91.86\% \\
random & \textbf{94.02\%}  & 89.61\% &  91.89\% \\
\bottomrule
\end{tabular*}
\vspace{2pt}
\end{table}

Section 4.2 introduces four types of edge features. To assess the impact of each feature, we carry out separate ablation studies on the TSPLIB benchmark dataset. Figure 8
illustrates the optimality gap variations under different feature ablation settings. The results indicate that all feature types contribute positively to the model performance. The KNN-based feature exhibits the most significant impact, consistent with its role as a widely used heuristic in TSP. The distance feature shows a relatively smaller effect, likely due to its inherent correlation with the KNN. In comparison, the Q-value feature, which acts as an additional refinement over KNN, has the least influence. The IMST feature, as a global edge feature, also plays an important role and shows a relatively large influence on performance.

\begin{figure}[htbp]
    \centering
    \includegraphics[width=0.8\linewidth]{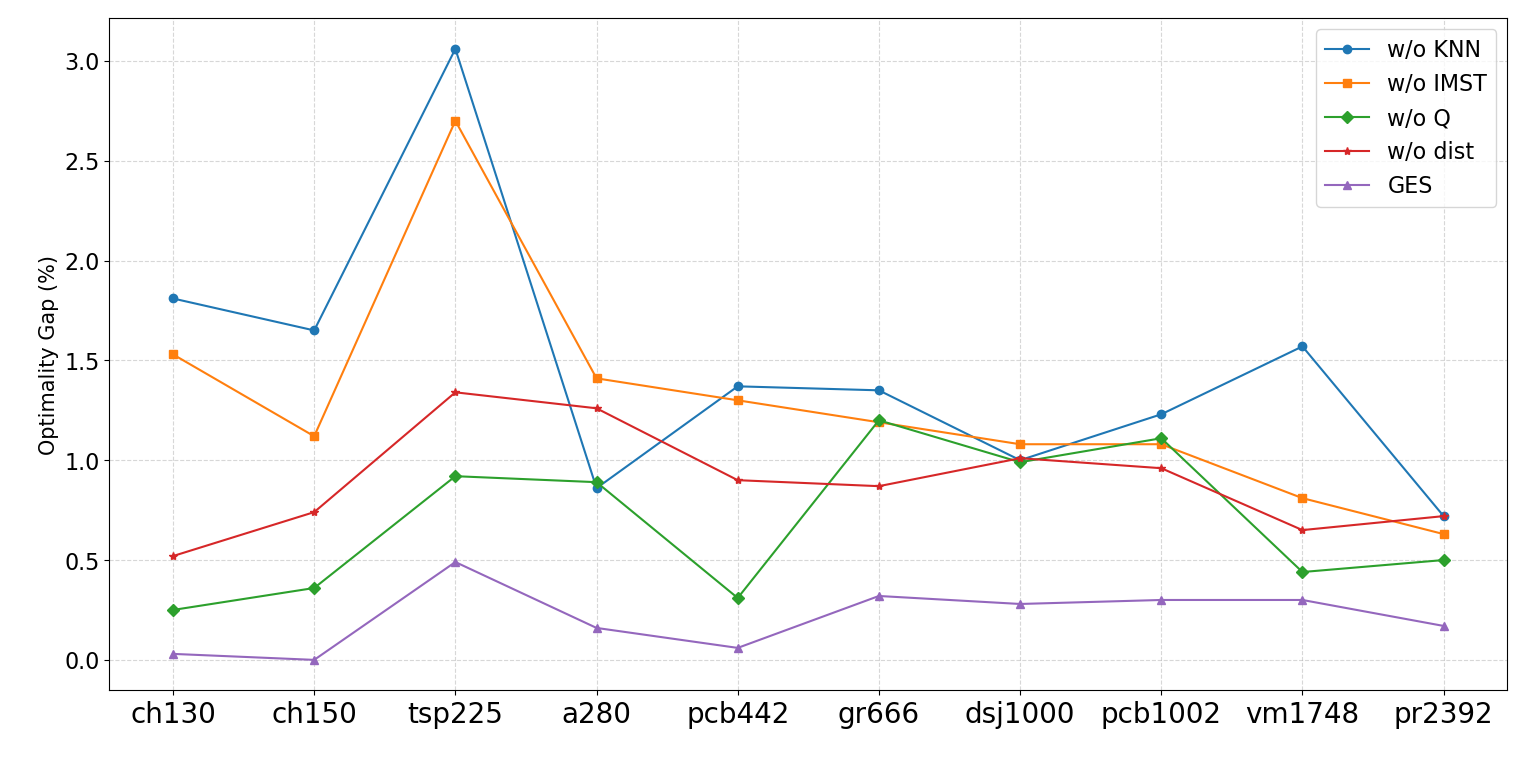}
    \caption{Ablation Results of Edge Features}
    \label{fig:wo}
\end{figure}

After selecting edges based on the threshold, we further incorporate the edges from the Christofides algorithm solution. As shown in Table 7 in appendices, without these additional edges, some instances may fail to admit a Hamiltonian cycle. In particular, for harder instances such as CLKhard and LKCChard, the number of infeasible cases increases significantly when Christofides-based augmentation is not applied.

\section{Conclusion}
In this paper, we propose GES, a learning-based graph sparsification framework for the Euclidean Traveling Salesman Problem. By combining geometric structures with data-driven learning, our method effectively reduces the size of the candidate edge set while preserving high-quality solutions.

Extensive experiments on the MATILDA and TSPLIB benchmarks demonstrate that GES consistently achieves superior performance compared to existing methods. Our approach attains higher pruning rates—often exceeding 95\% and even 99\% on large-scale instances, while maintaining an optimality gap within 1\%. Moreover, the results show that GES generalizes well across different datasets and significantly accelerates the solving process, especially for large-scale problems. However, our current framework is limited to solving the TSP and has not yet been extended to other combinatorial optimization problems.

Overall, this work highlights the effectiveness of integrating geometric heuristics with graph neural networks for combinatorial optimization. In future work, we plan to explore more advanced GNNs architectures and extensions to other combinatorial optimization problems.

\clearpage
\bibliographystyle{unsrt}
\bibliography{references}

@article{junger1995traveling,
  title={The traveling salesman problem},
  author={J{\"u}nger, Michael and Reinelt, Gerhard and Rinaldi, Giovanni},
  journal={Handbooks in operations research and management science},
  volume={7},
  pages={225--330},
  year={1995},
  publisher={Elsevier}
}

@article{spielman2011spectral,
  title={Spectral sparsification of graphs},
  author={Spielman, Daniel A and Teng, Shang-Hua},
  journal={SIAM Journal on Computing},
  volume={40},
  number={4},
  pages={981--1025},
  year={2011},
  publisher={SIAM}
}

@book{de2008computational,
  title={Computational geometry: algorithms and applications},
  author={De Berg, Mark and Cheong, Otfried and Van Kreveld, Marc and Overmars, Mark},
  year={2008},
  publisher={Springer}
}

@article{xu2020delaunay,
  title={Delaunay-triangulation-based variable neighborhood search to solve large-scale general colored traveling salesman problems},
  author={Xu, Xiangping and Li, Jun and Zhou, MengChu},
  journal={IEEE Transactions on Intelligent Transportation Systems},
  volume={22},
  number={3},
  pages={1583--1593},
  year={2020},
  publisher={IEEE}
}

@article{vinyals2015pointer,
  title={Pointer networks},
  author={Vinyals, Oriol and Fortunato, Meire and Jaitly, Navdeep},
  journal={Advances in neural information processing systems},
  volume={28},
  year={2015}
}

@article{kwon2020pomo,
  title={Pomo: Policy optimization with multiple optima for reinforcement learning},
  author={Kwon, Yeong-Dae and Choo, Jinho and Kim, Byoungjip and Yoon, Iljoo and Gwon, Youngjune and Min, Seungjai},
  journal={Advances in neural information processing systems},
  volume={33},
  pages={21188--21198},
  year={2020}
}

@article{joshi2019efficient,
  title={An efficient graph convolutional network technique for the travelling salesman problem},
  author={Joshi, Chaitanya K and Laurent, Thomas and Bresson, Xavier},
  journal={arXiv preprint arXiv:1906.01227},
  year={2019}
}

@article{bello2016neural,
  title={Neural combinatorial optimization with reinforcement learning},
  author={Bello, Irwan and Pham, Hieu and Le, Quoc V and Norouzi, Mohammad and Bengio, Samy},
  journal={arXiv preprint arXiv:1611.09940},
  year={2016}
}

@article{bresson2021transformer,
  title={The transformer network for the traveling salesman problem},
  author={Bresson, Xavier and Laurent, Thomas},
  journal={arXiv preprint arXiv:2103.03012},
  year={2021}
}

@inproceedings{fitzpatrick2021learning,
  title={Learning to sparsify travelling salesman problem instances},
  author={Fitzpatrick, James and Ajwani, Deepak and Carroll, Paula},
  booktitle={International Conference on Integration of Constraint Programming, Artificial Intelligence, and Operations Research},
  pages={410--426},
  year={2021},
  organization={Springer}
}

@article{xin2021neurolkh,
  title={Neurolkh: Combining deep learning model with lin-kernighan-helsgaun heuristic for solving the traveling salesman problem},
  author={Xin, Liang and Song, Wen and Cao, Zhiguang and Zhang, Jie},
  journal={Advances in Neural Information Processing Systems},
  volume={34},
  pages={7472--7483},
  year={2021}
}

@inproceedings{tian2024combhelper,
  title={Combhelper: a neural approach to reduce search space for graph combinatorial problems},
  author={Tian, Hao and Medya, Sourav and Ye, Wei},
  booktitle={Proceedings of the AAAI Conference on Artificial Intelligence},
  volume={38},
  number={18},
  pages={20812--20820},
  year={2024}
}

@article{hashemi2024comprehensive,
  title={A Comprehensive Survey on Graph Reduction: Sparsification, Coarsening, and Condensation},
  author={Hashemi, Mohammad and Gong, Shengbo and Ni, Juntong and Fan, Wenqi and Prakash, B Aditya and Jin, Wei},
  year={2024},
  publisher={Proceedings of IJCAI 2024}
}

@article{scarselli2008graph,
  title={The graph neural network model},
  author={Scarselli, Franco and Gori, Marco and Tsoi, Ah Chung and Hagenbuchner, Markus and Monfardini, Gabriele},
  journal={IEEE transactions on neural networks},
  volume={20},
  number={1},
  pages={61--80},
  year={2008},
  publisher={IEEE}
}

@article{velickovic2018graph,
  title={GRAPH ATTENTION NETWORKS},
  author={Velickovic, Petar and Cucurull, Guillem and Casanova, Arantxa and Romero, Adriana and Lio, Pietro and Bengio, Yoshua},
  journal={stat},
  volume={1050},
  pages={4},
  year={2018}
}

@inproceedings{christofides2022worst,
  title={Worst-case analysis of a new heuristic for the travelling salesman problem},
  author={Christofides, Nicos},
  booktitle={Operations Research Forum},
  volume={3},
  number={1},
  pages={20},
  year={2022},
  organization={Springer}
}

@article{achterberg2009scip,
  title={SCIP: solving constraint integer programs},
  author={Achterberg, Tobias},
  journal={Mathematical Programming Computation},
  volume={1},
  number={1},
  pages={1--41},
  year={2009},
  publisher={Springer}
}

@article{helsgaun2015solving,
  title={Solving the equality generalized traveling salesman problem using the Lin--Kernighan--Helsgaun algorithm},
  author={Helsgaun, Keld},
  journal={Mathematical Programming Computation},
  volume={7},
  number={3},
  pages={269--287},
  year={2015},
  publisher={Springer}
}

@article{hu2020open,
  title={Open graph benchmark: Datasets for machine learning on graphs},
  author={Hu, Weihua and Fey, Matthias and Zitnik, Marinka and Dong, Yuxiao and Ren, Hongyu and Liu, Bowen and Catasta, Michele and Leskovec, Jure},
  journal={Advances in neural information processing systems},
  volume={33},
  pages={22118--22133},
  year={2020}
}

@article{sun2020generalization,
  title={Generalization of machine learning for problem reduction: a case study on travelling salesman problems},
  author={Sun, Yuan and Ernst, Andreas and Li, Xiaodong and Weiner, Jake},
  journal={arXiv preprint arXiv:2005.05847},
  year={2020}
}
\clearpage
\appendix

\section{Technical Appendices and Supplementary Material}
This appendix supplements the line charts in the main text by reporting the corresponding detailed numerical results in tabular form.
\begin{table}[htbp]
\centering
\caption{Optimality Gap Comparison of TSP Solving on the TSPLIB }
\label{tab:Optimality Ratio}
\footnotesize
\begin{tabular*}{\textwidth}{@{\extracolsep{\fill}}lccc} 
\toprule
Statistic & SGN & Fitzpatrick & GES  \\
\midrule
ch130 & 1.05\% & 0.04\% & \textbf{0.03\%} \\
ch150 & 3.34\% & 0.02\% & \textbf{0.00\%} \\
tsp225 & 6.96\% & 0.87\% & \textbf{0.49\%} \\
a280 & 1.83\% & 0.98\% & \textbf{0.16\%} \\
pcb442 & 2.71\% & 0.57\% & \textbf{0.20\%} \\
gr666 & 3.57\% & 1.31\% & \textbf{0.32\%} \\
dsj1000 & 4.59\% & 1.20\% & \textbf{0.18\%} \\
pcb1002 & 4.97\% & 2.63\% & \textbf{0.30\%} \\
vm1748 & 5.29\% & 2.05\% & \textbf{0.30\%} \\
pr2392 & 4.81\% & 1.96\% & \textbf{0.10\%} \\
\bottomrule
\end{tabular*}
\end{table}

\begin{table}[htbp]
\centering
\caption{Pruning Rate Comparison of TSP Solving on the TSPLIB }
\label{tab:Optimality Ratio}
\footnotesize
\begin{tabular*}{\textwidth}{@{\extracolsep{\fill}}lccc} 
\toprule
Statistic & SGN & Fitzpatrick & GES  \\
\midrule
ch130 & 94.02\% & 92.08\% & \textbf{95.30\%} \\
ch150 & 95.97\% & 92.63\% & \textbf{96.00\%} \\
tsp225 & 96.53\% & 94.02\% & \textbf{97.42\%} \\
a280 & 97.88\% & 95.98\% & \textbf{97.90\%} \\
pcb442 & 97.11\% & 95.02\% & \textbf{98.64\%} \\
gr666 & 97.54\% & 95.95\% & \textbf{99.06\%} \\
dsj1000 & 98.03\% & 96.03\% & \textbf{99.37\%} \\
pcb1002 & 98.48\% & 96.03\% & \textbf{99.38\%} \\
vm1748 & 98.96\% & 96.38\% & \textbf{99.67\%} \\
pr2392 & 99.00\% & 96.52\% & \textbf{99.74\%} \\
\bottomrule
\end{tabular*}
\end{table}

\begin{table}[htbp]
\centering
\caption{Ablation Results of Edge Features }
\label{tab:Optimality Ratio}
\footnotesize
\begin{tabular*}{\textwidth}{@{\extracolsep{\fill}}lccccc} 
\toprule
Statistic & w/o KNN & w/o IMST & w/o Q & w/o dist & GES  \\
\midrule
ch130   & 1.81\% & 1.53\% & 0.25\% & 0.52\% & \textbf{0.03\%} \\
ch150   & 1.65\% & 1.12\% & 0.36\% & 0.74\% & \textbf{0.00\%} \\
tsp225  & 3.06\% & 2.70\% & 0.92\% & 1.34\% & \textbf{0.49\%} \\
a280    & 0.86\% & 1.41\% & 0.89\% & 1.26\% & \textbf{0.16\%} \\
pcb442  & 1.37\% & 1.30\% & 0.31\% & 0.90\% & \textbf{0.06\%} \\
gr666   & 1.35\% & 1.19\% & 1.20\% & 0.87\% & \textbf{0.32\%} \\
dsj1000 & 1.00\% & 1.08\% & 0.99\% & 1.01\% & \textbf{0.28\%} \\
pcb1002 & 1.23\% & 1.08\% & 1.11\% & 0.96\% & \textbf{0.30\%} \\
vm1748  & 1.57\% & 0.81\% & 0.44\% & 0.65\% & \textbf{0.30\%} \\
pr2392  & 0.72\% & 0.63\% & 0.50\% & 0.72\% & \textbf{0.17\%} \\
\bottomrule
\end{tabular*}
\end{table}
\clearpage

\begin{table}[htbp]
\centering
\caption{Impact of Christofides Edge Augmentation on Feasibility}
\label{tab:3/2}
\footnotesize
\begin{tabular*}{\textwidth}{@{\extracolsep{\fill}}lcc} 
\toprule
Statistic & GES & w/o Christofides  \\
\midrule
CLKeasy & 190/190 & 190/190  \\
CLKhard & 127/127 & 119/127  \\
LKCCeasy & 190/190 & 180/190 \\
LKCChard & 127/127 & 119/127  \\
easyCLK-hardLKCC & 190/190 & 187/190 \\
hardCLK-easyLKCC & 190/190 & 188/190 \\
random & 190/190 & 188/190  \\
\bottomrule
\end{tabular*}
\end{table}

\begin{figure}[htbp]
    \centering
    
    \begin{subfigure}{0.48\textwidth}
        \centering
        \includegraphics[width=\linewidth]{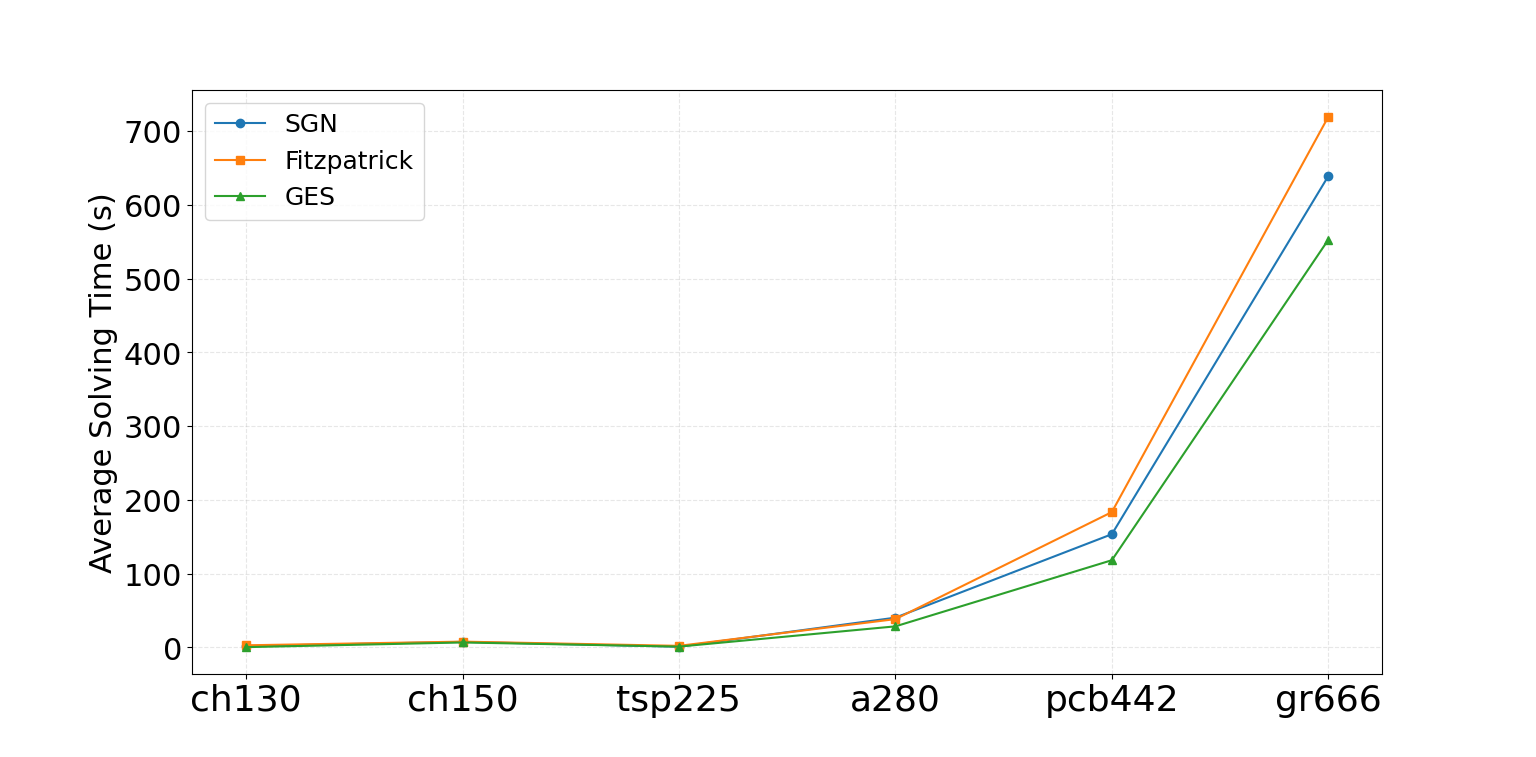}
        \caption{small-scale TSP instances}
    \end{subfigure}
    \hfill
    \begin{subfigure}{0.48\textwidth}
        \centering
        \includegraphics[width=\linewidth]{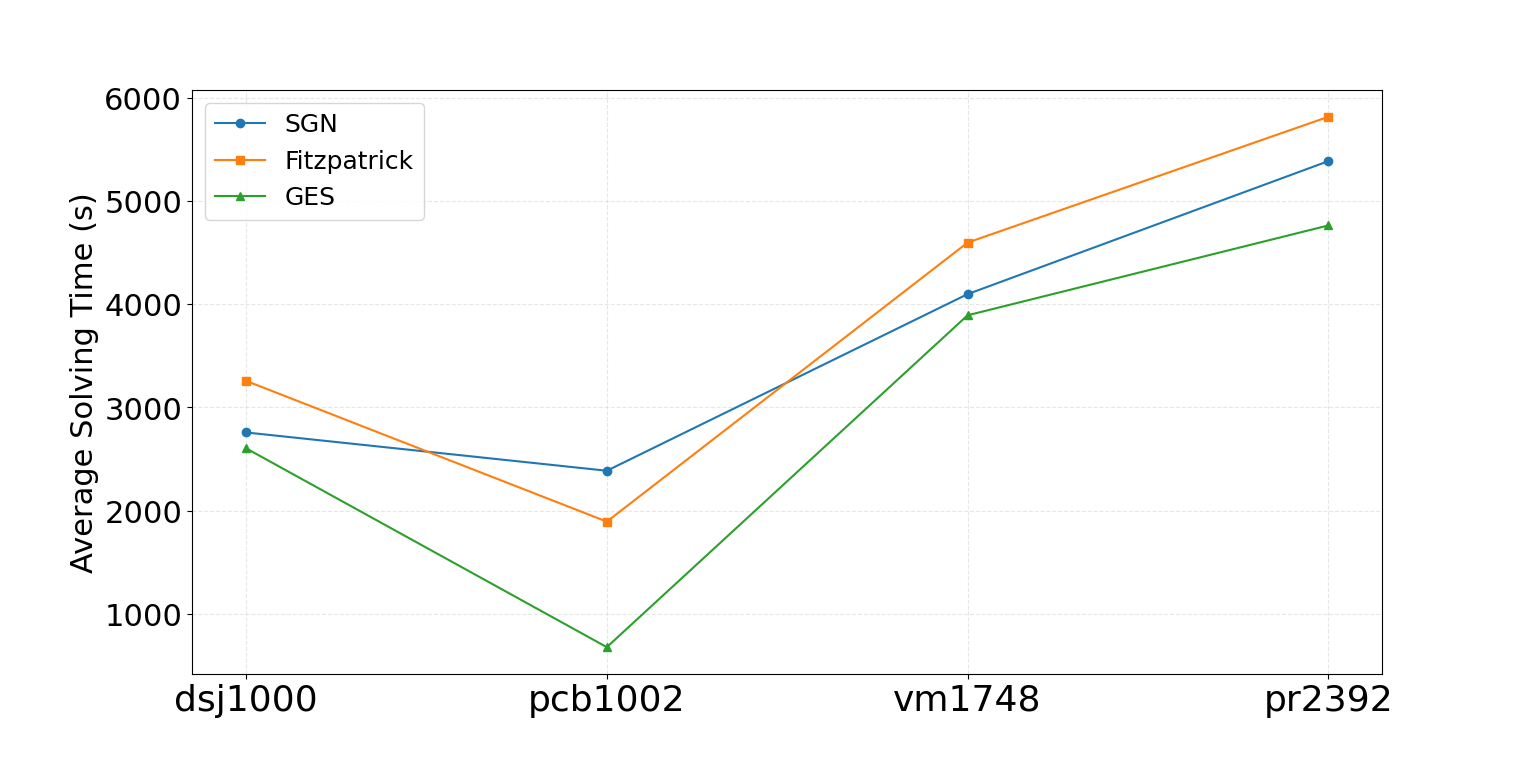}
        \caption{large-scale TSP instances}
    \end{subfigure}
    
    \caption{Comparison of running times on the TSPLIB}
    \label{fig:compare}
\end{figure}

\newpage
\clearpage

\end{document}